%% file: main.tex
\documentclass[11pt, letterpaper, copyright, logo]{googledeepmind}

\usepackage{multirow}
\usepackage[authoryear, sort&compress, round]{natbib}
\usepackage{tabularx}
\usepackage{enumitem}
\usepackage{lipsum}
\usepackage{amsmath}
\usepackage{pstricks, pst-node}
\usepackage{verbatim}
\usepackage{multirow}
\usepackage{multicol}
\usepackage{longtable}
\usepackage{array}
\usepackage{listings, listings-rust}
\usepackage{booktabs}
\usepackage{enumitem}
\usepackage{xspace}
\usepackage{bm}
\usepackage{bbm}
\usepackage{mathtools}
\usepackage{soul}
\usepackage{graphicx}
\usepackage[most, breakable, skins]{tcolorbox}
\usepackage{subcaption}
\usepackage{amssymb}
\usepackage{colortbl}
\usepackage{csquotes}
\usepackage[inkscapeformat=png]{svg}
\usepackage{tabularx,ragged2e}
\usepackage{makecell}
\usepackage{placeins}
\usepackage[dvipsnames]{xcolor}
\usepackage{hyperref}
\usepackage{pifont}
\usepackage[capitalise]{cleveref}
\usepackage{threeparttable}

\newcolumntype{L}[1]{>{\raggedright\let\newline\\\arraybackslash\hspace{0pt}}m{#1}}
\newcolumntype{C}[1]{>{\centering}m{#1}}

\newcolumntype{R}[1]{>{\raggedleft\let\newline\\\arraybackslash\hspace{0pt}}m{#1}}

\definecolor{ao}{rgb}{0.0, 0.0, 1.0}

\newcommand\vcent[1]{\vcenter{\hbox{#1}}}
\newcommand\loudspeaker[1][3]{\ensuremath{\vcent{\rule{.6ex}{.6ex}}\kern-.5ex%
  \vcent{\scalebox{.6}[1]{\rotatebox[origin=center]{90}{$\blacktriangle$}}}%
  \ifnum#1>0\relax\kern.05ex\vcent{\scalebox{.4}{\ttfamily)}}%
  \ifnum#1>1\relax\kern-.4ex\vcent{\scalebox{.56}{\ttfamily)}}%
  \ifnum#1>2\relax\kern-.55ex\vcent{\scalebox{.7}{\ttfamily)}}%
  \fi\fi\fi}%
}

\newcommand{\tick}{\ding{51}}

\let\cite\citep

\title{MedGemma 1.5 Technical Report}

\author[]{Google Research and Google DeepMind \footnote{\small{ See Contributions and Acknowledgments section for full author list. \\\hspace*{2.7em}\small{Corresponding authors: \{chufang, dangolden, asellerg\}@google.com}.}}}

\begin{document}

\begin{abstract}
\input{sections/abstract}  
\end{abstract}

\maketitle

\clearpage
\section{Introduction} 
\label{sec:intro}
\input{sections/intro}
\section{Methods}
\label{sec:methods}
\input{sections/methods}

\section{Discussion}
\label{sec:discussion}
\input{sections/discussion}

\section{Conclusion}
\label{sec:conclusion}
\input{sections/conclusion}

\section{Model availability}
The models have been released openly at the main Google Health AI Developer Foundations site at \mbox{\url{https://goo.gle/hai-def}}. Further details specifically about the MedGemma collection of models can be found at \mbox{\url{https://goo.gle/medgemma}}.

\newpage
\section{Contributions and Acknowledgments}
\input{sections/ack}

\clearpage
\bibliography{main}

\newpage
\appendix
\label{sec:appendix}

\input{sections/appendix}


\end{document}

%% file: sections/abstract.tex
We introduce MedGemma 1.5 4B, the latest model in the MedGemma collection. MedGemma 1.5 expands on MedGemma 1 by integrating additional capabilities: high-dimensional medical imaging (CT/MRI volumes and histopathology whole slide images), anatomical localization via bounding boxes, multi-timepoint chest X-ray analysis, and improved medical document understanding (lab reports, electronic health records). We detail the innovations required to enable these modalities within a single architecture, including new training data, long-context 3D volume slicing, and whole-slide pathology sampling. Compared to MedGemma 1 4B, MedGemma 1.5 4B demonstrates significant gains in these new areas, improving 3D MRI condition classification accuracy by 11\% and 3D CT condition classification by 3\% (absolute improvements). In whole slide pathology imaging, MedGemma 1.5 4B achieves a 47\% macro F1 gain. Additionally, it improves anatomical localization with a 35\% increase in Intersection over Union on chest X-rays and achieves a 4\% macro accuracy for longitudinal (multi-timepoint) chest x-ray analysis. Beyond its improved multimodal performance over MedGemma 1, MedGemma 1.5 improves on text-based clinical knowledge and reasoning, improving by 5\% on MedQA accuracy and 22\% on EHRQA accuracy. It also achieves an average of 18\% macro F1 on 4 different lab report information extraction datasets (EHR Datasets 2, 3, 4, and Mendeley Clinical Laboratory Test Reports). Taken together, MedGemma 1.5 serves as a robust, open resource for the community, designed as an improved foundation on which developers can create the next generation of medical AI systems. Resources and tutorials for building upon MedGemma 1.5 can be found at \url{https://goo.gle/medgemma}.


%% file: sections/intro.tex
Expanding the capabilities of open-weight medical foundation models to encompass complex, high-dimensional modalities is essential for comprehensive healthcare AI development. While recent advancements \cite{yang2024advancing, sellergren2025medgemma} have demonstrated the utility of multimodal models in standard 2D imaging tasks, the number of models and benchmarks addressing more complicated imaging tasks are more limited. In this technical report, we introduce MedGemma 1.5, an updated model in the MedGemma collection integrating support for high-dimensional, volumetric, and longitudinal data, along with existing support for 2D imaging and text-based knowledge and reasoning, all within a single unified architecture.

Specifically, MedGemma 1.5  expands the multimodal capabilities of previous releases by introducing
native support for four medical imaging capabilities including:  (1) 3D radiology interpretation (both CT and MRI volumes),
(2) whole slide image (WSI) interepretaion for histopathology,
(3) fine-grained anatomical localization for X-rays via bounding boxes, and (4) multi-timepoint radiology analysis. 
Through additional curated training datasets, we also incorporated improved capabilities for medical document (PDF) understanding and improved text-based clinical reasoning. As the first open model to achieve these diverse baseline capabilities in a single architecture, MedGemma 1.5 serves as a robust, open resource for the community, designed as an improved foundation on which developers can create the next generation of medical AI systems.

\begin{figure}[htp] \centering \includegraphics[width=1\linewidth]{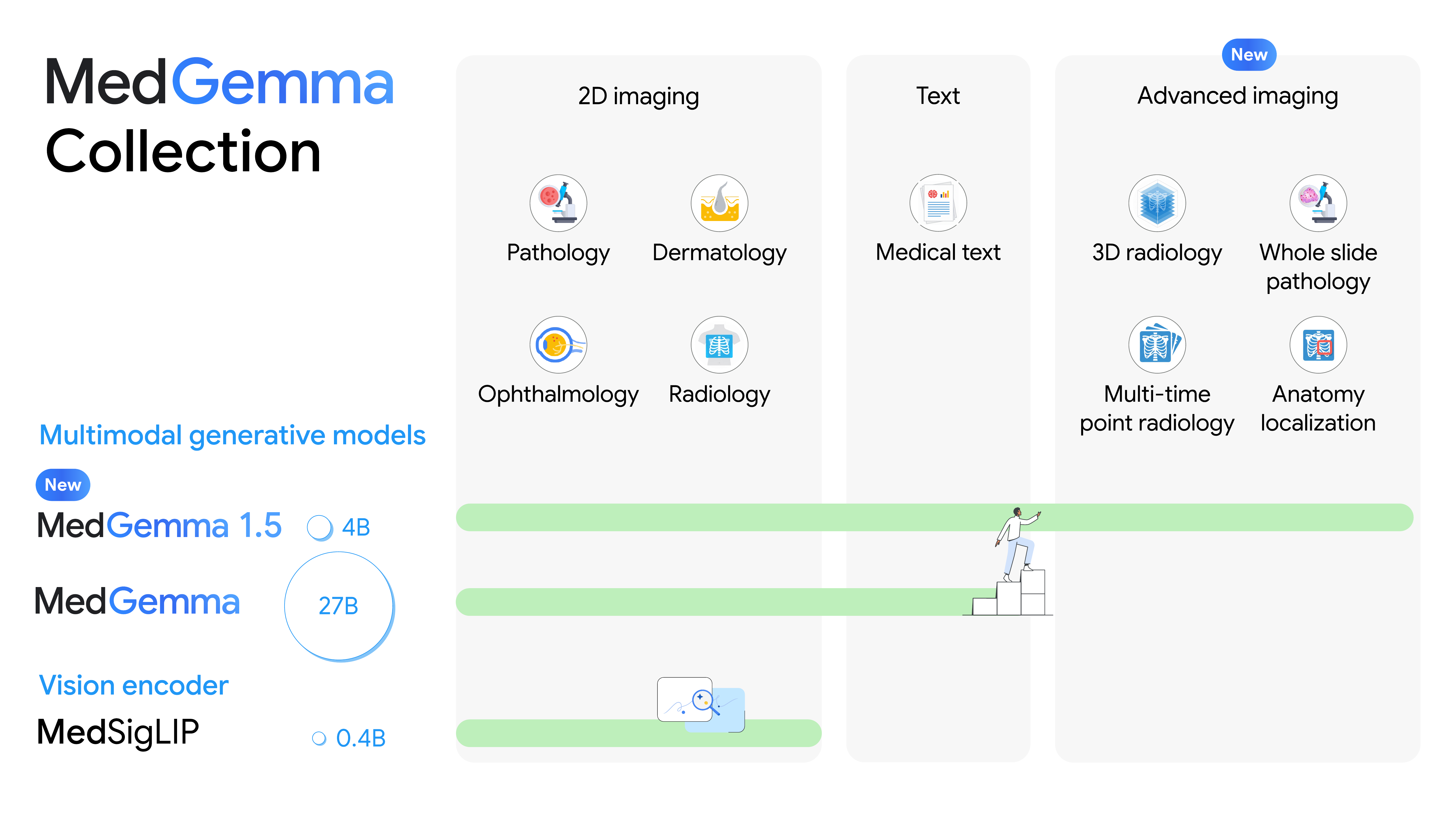} \caption{\small{\textbf{Overview of model capabilities within the MedGemma collection.} The updated MedGemma 1.5 4B architecture now supports 3D radiology (CT/MRI volumes), pathology whole slide imaging (WSI), anatomical localization, and multi-timepoint analysis. The original MedGemma 27B model remains available for complex clinical knowledge and reasoning tasks and MedSigLIP remains available for medical image classification and retrieval tasks.}} \label{fig:MedGemma_1_5_overview} \end{figure}

\begin{figure}[thbp]
    \centering
    \includegraphics[width=\linewidth]{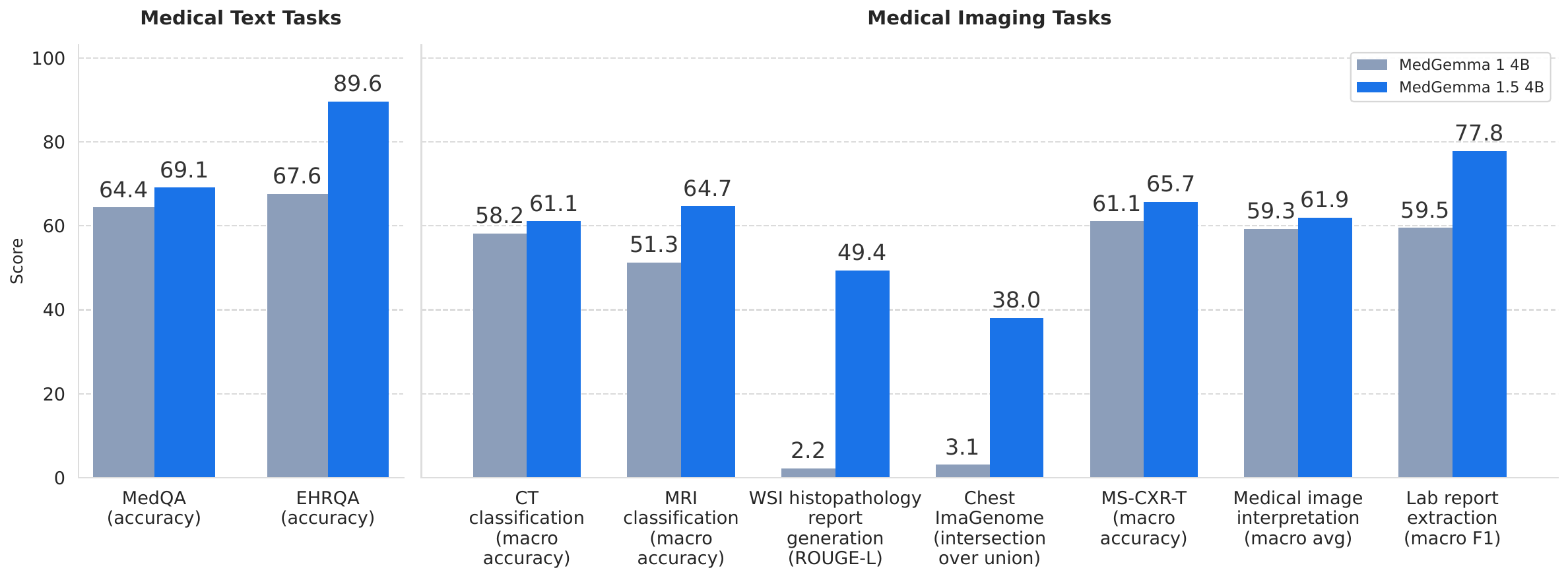}
    \caption{The left panel details accuracy on medical text Q\&A tasks (MedQA and EHRQA), while the right panel highlights performance across diverse medical imaging capabilities. Notably, the "medical image interpretation" score represents an unweighted macro average of the model's performance across 7 distinct imaging tasks including MIMIC-CXR (both RadGraph F1 and report generation macro F1), CheXpert (unweighted average across 5 conditions), CXR 14 (unweighted average across 3 conditions), Path MCQA, DermMCQA, and EyePACS. "Lab report extraction" is macro-averaged over results from EHR dataset 2, 3 and 4 as well as Mendeley Clinical Laboratory Test Reports (macro F1). All scores are reported as percentages. While the out-of-the-box performance is highly promising, the model is not meant to be deployed without the necessary clinical fine-tuning. Fine-tuning may improve results and adapt the framework for practical use. \label{fig:combined_medical_tasks_shared}}
\end{figure}

%% file: sections/methods.tex
Similar to MedGemma 1, MedGemma 4B 1.5 is based off of Gemma3 \cite{gemmateam2025gemma3technicalreport} with the same architecture. The vision encoder is 400M MedSigLIP encoder \cite{sellergren2025medgemma}. 

The methodological updates for MedGemma 4B version 1.5 primarily fall into three categories: first, we incorporated several new training datasets with the goal of broadening medical knowledge and expanding capabilities for ``high-dimensional'' medical imaging including 3D CT and MRI volumes and Whole Slide Images (WSIs) for histopathology. Second, we implemented minor modifications to the modeling methodology to improve training efficiency and model capabilities. Third, we expanded evaluations to assess model performance on updated capabilities.


\begin{table}[t]

\centering
\caption{\small{\textbf{Additional training data for MedGemma 4B 1.5 relative to MedGemma 1}}}
\footnotesize
\label{tab:1.5_training-dataset-info}
\resizebox{\textwidth}{!}{
\renewcommand{\arraystretch}{1.25}
\begin{tabular}{l|l|r|l|p{2.6in}}
\Xhline{2.5\arrayrulewidth}
\textbf{Modality} & \textbf{Dataset}    & \textbf{No. Train Examples} & \textbf{Training stages} & \textbf{Description} \\
\Xhline{2\arrayrulewidth}

                          

\multirow{9}{*}{Radiology} & 
CXR-IND1     & 605,732 & PT, Distill, RL & Dataset of chest X-ray images and free text reports from a large hospital system based in India \\
& CT Dataset 1  & 282,963    & PT, Distill, RL  & Dataset of different axial CT studies across body parts (head, chest, abdomen) from a US-based radiology outpatient diagnostic center network. \\

& MRI Dataset 1 & 167,674   & PT, Distill, RL  & Dataset of different axial multi-parametric MRI studies across body parts (head, abdomen, knee) from a US-based radiology outpatient diagnostic center network. \\

& Chest ImaGenome \cite{wu2021chest} & 39,968  & RL  & Dataset of chest X-ray with sequential images and bounding boxes from IBM Research. \\
\hline
\multirow{3}{*}{Pathology Whole Slide Imaging} & Internal WSI Histopathology    &  335,825  & PT, RL & WSIs with paired final diagnosis text reports. WSIs were tiled into individual patches and aggregated for input as described in the Methods. \\ 

\hline
\multirow{5}{*}{Dermatology} & Dermatology Dataset 4    & 25,560 & PT, Distill & Dermatology dataset featuring multiple images and longitudinal visits and records from Japan. \\


& Dermatology Dataset 5    & 87,879 & PT, Distill, RL & Dermatology dataset featuring unlabeled images. \\


& ISIC     & 40,269 & PT, Distill & Dermoscopic images with lesion diagnoses or attribute labels.\\

\hline
\multirow{12}{*}
{\makecell[l]{Electronic Health Record\\and Laboratory Reports}} & EHRQA$^*$     & 9,809 (QA Pairs)  & Distill  & Question/answer dataset drawn from synthetic FHIR records. 
\\

& EHR Dataset 2    & 1,539 (2,846 pages) & Distill & Lab Reports across different departments in histopathology such as Biochemistry, Clinical histopathology, Hematology, Microbiology and Serology \\
                           
& EHR Dataset 3  & 6214 (18,035 pages) & Distill  & Lab Reports across different departments in histopathology such as Biochemistry, Clinical histopathology, Hematology, Microbiology and Serology 
\\
& EHR Dataset 4  & 497 (1,278 pages) & Distill  & Synthetic reports based on a Latex powered custom PDF generator for different lab report templates in the US \\
                             
& EHR Dataset 5 & 33,882 (user queries)  & Distill  & Synthetic dataset of approximately 60,000 health-relevant user queries \\

\Xhline{2.5\arrayrulewidth}
\end{tabular}
}
\begin{tablenotes}
  \scriptsize 
  \item $^*$Note that EHRQA was previously not included in the training of MedGemma 1 4B but was included in training data MedGemma 1 27B. (This dataset is also referred to as EHR Dataset 1 in the Model Card.)
  \item PT: Continued Pretraining (supervised finetuning of LLM), RL: Reinforcement learning. Distill: Distilled from teacher model(s).
\end{tablenotes}
\end{table}

\begin{table}[htb]
\centering
\footnotesize
\caption{\small{\textbf{Additional evaluation datasets for MedGemma 1.5}}}
\label{tab:evals_data_1.5}
\resizebox{\textwidth}{!}{
\renewcommand{\arraystretch}{1.15}
\begin{tabular}{@{}l|p{2.6in}|l|r|c|c}
\Xhline{2.5\arrayrulewidth}
\textbf{Task}  & \textbf{Dataset}& \textbf{Modality}  & \textbf{No. Eval. Examples} & \textbf{OOD\textsuperscript{$\dagger$}} & \textbf{Public\textsuperscript{*}}                 \\
\Xhline{2\arrayrulewidth}
\multirow{1}{*}{Medical text QA}

& EHRNoteQA \cite{kweon2024ehrnoteqa}            & Text & 962   & \tick & \tick   \\

\hline
\multirow{1}{*}{CXR temporal analysis}
& MS-CXR-T \cite{bannur2023learning} & Radiology longitudinal   & 1326    & - & \tick \\ 

\hline

\multirow{2}{*}{3D CT classification}
& CT Dataset 1        & Radiology CT (Volumes) & 1229  & - & - \\ 
& CT-RATE \cite{hamamci2024developing}       & Radiology CT (Volumes) & 1558  & \tick & \tick  \\
\hline
\multirow{1}{*}{CXR Finding localization}
& Chest ImaGenome \cite{wu2021chest}        & Radiology localization      & 10000  & - & \tick \\
\hline
\multirow{1}{*}{Pathology WSI to text}
& WSI Histopath 
& Pathology WSIs & 9614  & - & -     \\
\hline
\multirow{5}{*}{Document Understanding}
& EHR Dataset 2 & PDF documents & 170 (304 pages)  & - & -     \\
& EHR Dataset 3 & PDF documents & 702 (2005 pages)  & - & -     \\
& EHR Dataset 4 & PDF documents & 56 (143 pages)  & - & -     \\
& Mendeley Clinical Laboratory Test Reports \cite{abdelmaksoud2022clinical} & PNG images & 14  & - & \tick     \\
\Xhline{2.5\arrayrulewidth}            
\end{tabular}
}
\begin{tablenotes}
  \scriptsize 
  \item $\dagger$ Out of Distribution: Data not seen during any model development stages.
  \item \textsuperscript{*} Denotes whether a dataset is publicly available or an internal dataset.
\end{tablenotes}

\end{table}

\subsection{Pretraining} For training MedGemma 1.5, the vision encoder of MedGemma 1 (MedSigLIP) \cite{sellergren2025medgemma, zhai2023sigmoid} was frozen, and the language decoder underwent additional pretraining (supervised finetuning of the LLM) to build upon the initial baseline capabilities. We incorporated both the text and interleaved imaging data from the original Gemma mixture as well as the new medical domain image-text paired data as summarized in Table \ref{tab:1.5_training-dataset-info}. 
These datasets were utilized via a combination of additional pretraining, distillation, and reinforcement learning as indicated in the table and described below. 


We added new modalities for radiology and new data for dermatology and histopathology, and EHR/lab report understanding. Specifically, to train MedGemma 1.5 4B, the internal dermatology dataset was expanded to include additional examples from the open (CC-0) components of the ISIC 2017 and 2018 datasets \cite{gutman2016skin, codella2018skinlesionanalysismelanoma, Tschandl_2018} as well as an internal set of images from a large hospital in Japan (adding to the dermatology datasets used previously and  described in \Citet{yang2024advancing}). Additionally, internal radiography datasets--CXR-IND1, CT Dataset 1, and MRI Dataset 1--were also added to pretraining.

\subsection{Post-training} The knowledge acquired from pretraining was refined in the post-training stage through a combination of distillation and reinforcement learning (RL), and we used the same recipes as Gemma 3, but with additional medical data for both steps. Distillation here means we sample 256 teacher model logits per token, weighted by teacher probabilities. The student
learns the teacher’s distribution within these samples via cross-entropy loss \cite{gemmateam2025gemma3technicalreport}. 

The distillation process for MedGemma 1.5 was augmented by incorporating additional domain-specific teacher models in addition to an improved large instruction-tuned (IT) teacher. For example, to enhance high-dimensional medical imaging capabilities, we trained supplementary teachers on CT Dataset 1, MRI-Dataset 1, and Internal histopathology. 
As indicated in Table~\ref{tab:1.5_training-dataset-info}, additional Distill and/or RL training was applied across multiple modalities--including radiology, dermatology, and whole slide pathology imaging--to further align the model's performance with clinical visual tasks.
To enabled improved document understanding, we distilled on EHRQA (synthetic records created by Synthea \cite{walonoski2018synthea}.) as well as EHR datasets 2, 3, 4, and 5.

\subsection{Preprocessing}
\subsubsection{Preprocessing of Volumetric Images}
\label{sec:3d_ct}

Since the image encoder can only process 2D RGB images, 3D~CT and MR image volumes were preprocessed to sequences of individual 2D~axial images, each of which was rescaled to the image encoder's input dimension of of~$896 \times 896$ pixels.

We capped the number of axial slices per query to a maximum of 85 during training and evaluation (which amount to 21,760 vision tokens), to stay below a total of 32K tokens when including the radiology report indication (included in the prompt) and the findings (target), in order to keep the memory requirements during training manageable. Slices could originate from multiple z-stacked volumes per study. Inclusion criteria for each of these volumes were: a) a maximum of $512~\times 512$ pixels per slice, b) axial orientation, c) slices with the same thickness, and d) at least five slices. For CT studies, these included volumes with different reconstruction kernels originating from the same scan, and for MRI studies volumes representing different sequences and contrasts, including T1-weighted (T1w), T2-weighted (T2w), GRE, and SWI. For computational efficiency, if the final z-stacked volume had more then 85 slices in total, we sampled slices equidistantly across the z-axis (of the stacked volume). 

Regarding the mapping of voxel values, as with our previous work \citep{sellergren2025medgemma} for single slice CT images, we employed multi-channel windowing to map raw Hounsfield Units (HU) to RGB values, since the image encoder was trained only with 256 different intensity values per channel. Specifically, we used the following CT window mappings:
\begin{itemize}
    \item Red Channel (-1024, 1024): Given the heterogeneity of our training data (spanning brain, chest, and abdomen), this wide window ensures that morphological boundaries, from air-filled lung parenchyma to dense cortical bone, remain visible across all anatomical regions.

    \item Green Channel (-135, 215): Since the Green channel contributes most significantly to luminance in standard image processing, we mapped the complex soft-tissue data here to leverage the encoder's sensitivity to texture in visceral organs and mediastinal structures.

    \item Blue Channel (0, 80): This narrow, high-contrast window highlights subtle attenuations in brain parenchyma (gray/white matter differentiation) and acute intracranial hemorrhages, as well as vascular calcifications.
\end{itemize}

By contrast, since voxel values of MR images are all relative and no physiological windows exist, no windowing was applied, similar to the SigLip encoder \cite{sellergren2025medgemma}. Instead we normalized them per volume using min-max normalization, and set the R, G, and B channels to the same value.

\subsubsection{Preprocessing of Histopathology Whole-slide Images} 
\label{sec:wsi}
The histopathology WSI preprocessing pipeline was designed to convert WSIs into a sequence of representative tissue patches suitable for multi-modal learning. To ensure computational efficiency and relevance, we restricted patch extraction to tissue-containing regions. A tissue mask was generated for each slide at a low resolution (5x magnification). We employed a custom multi-stage tissue segmentation algorithm \cite{ahmed2025polypath} operating in the HSV (Hue, Saturation, Value) color space.


For each WSI, a single optical magnification level was stochastically selected for patch extraction, with probabilities set to approximate uniformity across standard diagnostic levels: P(5x)=0.34, P(10x)=0.33, and P(20x)=0.33.
The high-resolution tissue mask was downscaled to match the target extraction stride.
Non-overlapping patches of size 896 by 896 pixels were extracted from a regular grid defined by the tissue mask, with the stride equal to the patch size.
To maintain a fixed sequence length for downstream models, we enforced a cap of 126 patches per slide (leading to 32256 vision tokens) via random subsampling without replacement. Note that the number of images we used for long vision tasks differ for CT and WSI due to having different-length text inputs alongside them (i.e. CT had longer text inputs compared to WSI). Crucially, the original spatial ordering of the selected patches was preserved to retain relative positional context. The resulting patches were encoded as PNG images and stored alongside the slide caption. We chose a higher number of images per query than for CT and MRI examples above (126 versus 85), since fewer tokens were needed to encode image captions.

This processing is applied to an internal pathology dataset with a total of $\sim$335,825 whole slide images-text pairs for pretraining, distillation, and RL (on token-level ROUGE-L) as well as the 9,614 pairs used for evaluation.



\section{Evaluations}

Results of evaluations for original tasks from \citet{sellergren2025medgemma} are shown in Table~\ref{tab:1.5_vs_1.0_original_task_eval} and results of evaluations on new tasks are shown in Table~\ref{tab:new_datasets_eval_extended} and Figure~\ref{fig:combined_medical_tasks_shared}. 

\begin{table}[thbp]
\centering
\caption{\textbf{MedGemma 1.5 performance for original MedGemma 1 evaluation tasks} We bolded best performing small models and large models separately. \label{tab:1.5_vs_1.0_original_task_eval}}
\resizebox{\linewidth}{!}{
\begin{threeparttable} 
\begin{tabular}{l@{}l l |c c c c | c c}
\toprule
\multicolumn{3}{l|}{} & \multicolumn{4}{c|}{\textbf{Small Models}} & \multicolumn{2}{c}{\textbf{Large Models}} \\

    &Task & Metric & \makecell[c]{MedGemma\\1 4B} & \makecell[c]{MedGemma\\1.5 4B}  & \makecell[c]{MedGemma\\1 27B}  & \makecell[c]{Qwen3 VL\\4B} & \makecell[c]{Gemini 3\\Flash} & \makecell[c]{Gemini 3\\Pro}\\
\midrule
    &\textbf{Text evaluation}&&&  & & & & \\
\midrule
    &MedQA (4-op) \cite{medqa} & Accuracy & 64.4 & 69.1  & \textbf{85.3} & 76.8 & 94.3 & \textbf{95.1} \\
    &MedMCQA \cite{medmcqa}&  Accuracy & 55.7 & 59.8  & \textbf{70.2} & 63.7 & 85.2 & \textbf{86.1} \\
    &PubMedQA \cite{pubmedqa}&  Accuracy & 73.4 & 67.6  & \textbf{77.2} & 74.8 & 80.8 & \textbf{82.2} \\
    &MMLU Med \cite{mmlu} & Accuracy & 70.0 & 69.7  & \textbf{86.2} &  78.3 & 87.5 & \textbf{88.1} \\
    &MedXpertQA$^*$ (Text Only) \cite{zuo2025medxpertqa} & Accuracy & 14.2 & 16.4  & \textbf{21.8} & 17.5 & 67.7 & \textbf{78.2} \\
    &AfriMed-QA \cite{afrimed}&  Accuracy & 52.0 & 56.0  & \textbf{72.0} & 68.0 & \textbf{88.0} & 76.0 \\
\midrule
    &\multicolumn{2}{@{}l|}{\textbf{Electronic health record information retrieval}} & & & & \\
\midrule
    &EHRQA & Accuracy & 67.6 & 89.6  &\textbf{ 90.5} & 87.3& 94.9 & \textbf{95.2} \\
\midrule
    &\multicolumn{2}{@{}l|}{\textbf{Medical image classification}} & & & & \\
\midrule
    &MIMIC CXR$^\diamond$ (Med-Gemini Test Set) & Average F1 (5 conditions) & 88.9 & 89.5  & \textbf{90.0} & 78.5 & \textbf{88.4} & 85.0 \\
    &MIMIC CXR$^\diamond$ (MAIRA test set) & Average F1 (5 conditions) & 40.5 & \textbf{41.5}  & 40.2& 31.1 & 37.8 & \textbf{39.4 }\\
    &CheXpert CXR \cite{irvin2019chexpert} & Average F1 (5 conditions) & 48.1 & 48.2  & \textbf{49.9} & 33.5 & 47.8 & \textbf{51.2} \\
    &CXR14 \cite{majkowska2020chest} & Average F1 (3 conditions) & \textbf{50.1} & 48.4  & 45.3& 34.6 & 44.8 & \textbf{46.9} \\
    &DermMCQA \cite{liu2020deep} & Accuracy & 71.8 & \textbf{73.5}  & 71.7& 68.0 & 79.5 & \textbf{84.0} \\
    &PathMCQA \cite{sellergren2025medgemma}  & Accuracy & 69.8 & 70.0  & \textbf{71.6} & 41.8 & \textbf{59.3} & 59.1 \\
    &EyePACS \cite{cuadros2009eyepacs} & Accuracy & 64.9 & \textbf{76.8}  & 75.3 & 41.9 & \textbf{66.9} & 63.4 \\
\midrule
    &\multicolumn{2}{@{}l|}{\textbf{Visual question answering}} & & & & \\
\midrule
    &SlakeVQA \cite{liu2021slake} & Tokenized F1 & \textbf{72.3} & 59.8  & 70.3 & 53.5 & 62.5 & \textbf{62.8} \\
    &VQA-RAD \cite{lau2018dataset} & Tokenized F1 & \textbf{49.9} & 48.1  & 46.7 & 46.9 & 59.5 & \textbf{60.9} \\
\midrule
    &\multicolumn{2}{@{}l|}{\textbf{Knowledge and reasoning}} & & & & \\
\midrule
    &MedXpertQA$^*$(text + MM) \cite{zuo2025medxpertqa}  & Accuracy & 18.8 & 26.4  & \textbf{26.8} & 21.9 &  72.8 & \textbf{74.4} \\
\midrule
    &\multicolumn{2}{@{}l|}{\textbf{Report generation}} & & & & \\
\midrule
    &MIMIC CXR & Radgraph F1 \cite{jain2021radgraph} & 21.9 & \textbf{27.2}  & 27.0 & - & 7.4 & \textbf{20.8} \\
\bottomrule
\end{tabular}
\begin{tablenotes}
\item $^*$Indicates out of distribution for MedGemma

\item $^\diamond$MIMIC CXR \cite{johnson2019mimic} (1) Med-Gemini test set: radiologist-adjudicated labels with missing and uncertain labels excluded~\cite{yang2024advancing} and (2) MAIRA test set: labels from~\citet{MIMIC-CXR-JPG2019} with missing and uncertain labels considered to be negative, with~\citet{hyland2023maira} test set.
\end{tablenotes}
\end{threeparttable}
}
\end{table}

\begin{table}[thbp]
\centering
\caption{\small\textbf{New Evaluation Task Results}}
\label{tab:new_datasets_eval_extended}
\resizebox{\linewidth}{!}{
\begin{tabular}{@{}l l |c c c c c c | c c |c}
\toprule
\multicolumn{2}{l|}{} & \multicolumn{6}{c|}{\textbf{Small Models}} & \multicolumn{2}{c|}{\textbf{Large Models}} &  \\
Task & Metric & \makecell[c]{MedGemma\\1 4B} & \makecell[c]{MedGemma\\1.5 4B} & \makecell[c]{MedGemma\\1 27B} & \makecell[c]{Qwen3\\VL 4B} & \makecell[c]{Gemma 3\\4B} & \makecell[c]{Gemma 3\\27B} & \makecell[c]{Gemini 3.0\\Flash} & \makecell[c]{Gemini 3.0\\Pro} & \makecell[c]{External\\SOTA}  \\
\midrule
\multicolumn{2}{@{}l|}{\textbf{Text Only}} &&&&&&&&& \\
\midrule
EHRNoteQA & Accuracy & 79.4 & 80.4 &\textbf{ 90.7} & 90.6 & 78.0 & 90.3 & 93.9 & \textbf{95.0} & 95.15-97.16\textsuperscript{(1)} \\
\midrule
\multicolumn{2}{@{}l|}{\textbf{Document Understanding}} &&&&&&&&& \\
\midrule
EHR Dataset 2 & Macro F1 & 78 & 91 & 76 & - & 84 & \textbf{93} & 92 & \textbf{93} & - \\
EHR Dataset 3 & Macro F1 & 50 & 71 & 66 & - & 61 & \textbf{74} & \textbf{74} & \textbf{90} & -\\
EHR Dataset 4 & Macro F1 & 25 & \textbf{64} & 5 & - & 41 & 52 & \textbf{82} & 81 & -\\
\makecell[l]{Mendeley Clinical\\Laboratory Test Reports}  & Macro F1 & 85 & 85 & 69 & - & 83 & \textbf{89} & 89 & \textbf{90} & -\\
\midrule
\multicolumn{2}{@{}l|}{\textbf{CXR Analysis}} &&&&&&&&& \\
\midrule
\makecell[l]{Chest ImaGenome\\(Localization)} & Mean IoU & 3.1 & \textbf{38.0} & 16.0 & 8.7 & 5.7 & 8.1 & 38.5 & \textbf{39.1} & 30.7-34.4 \textsuperscript{(2)}\\
\midrule
MS-CXR-T$^\dagger$ (Temporal) & Macro-Accuracy & 61.1 & \textbf{65.7} & 50.1 & 53.5 & 59.0 & 52.7 & \textbf{67.3} & 62.9 & 68.5\textsuperscript{(3)}\\
\midrule
\multicolumn{2}{@{}l|}{\textbf{High-Dimension Image Analysis}} &&&&&&&&& \\
\midrule
CT Dataset 1 (3D CT) & Accuracy & 58.2 & \textbf{61.1} & 57.8 & 52.8 & 54.5 & 55.7 & \textbf{62.9} & 61.0 & -\\
MRI Dataset 1 (3D MRI) & Accuracy & 51.3 & \textbf{64.7} & 57.4 & 49.6 & 51.1 & 50.5 & \textbf{60.3} & 55.5 & -\\
WSI Histopath & ROUGE-L & 2.2 & \textbf{49.4} & 4.1 & ? & 2.3 & 3.2 & \textbf{13.9} & 12.2 & 49.8\textsuperscript{(4)} \\
\bottomrule
\end{tabular}
}
\begin{tablenotes}[itemsep=]\scriptsize
    \item $^\dagger$ During pretraining, mentions of temporal relationships in CXR reports were removed.
    \item \textsuperscript{(1)} GPT-4 \cite{kweon2024ehrnoteqa}. A range is given since original results are reported separately for levels 1 and 2.
    \item \textsuperscript{(2)} CoCa-CXR \cite{chen2025coca}. A range is given since original results are reported separately for current and prior images.
    \item \textsuperscript{(3)} BioViL-T \cite{bannur2023learning}
    \item \textsuperscript{(4)} PolyPath \cite{ahmed2025polypath}
\end{tablenotes}
\end{table}

Unless reported otherwise, all evaluations that we performed consisted of a single inference run per example. For MedGemma 1.5 evaluations, a temperature of 0.0 was used on all evaluations based on informal measurements of tune set performance. For other baseline models, the default temperature was used. Temperature 0.0 was kept for new dataset evaluations of MedGemma 1 for consistency. For evaluations of all other models, on all datasets, each model’s default temperature and top-k were used. Note that results may vary with other choices of temperature and top-k. All evaluations were conducted using data sources or splits of data sources that were completely held out from MedGemma training. Prompts were updated for MedGemma 1.5 as summarized in Appendix~\ref{sec:medgemma1.5_prompts}. Note that changes to prompts sometimes had a significant effect on benchmark performance and further optimization of prompts is likely possible.

\subsection{Existing MedGemma Benchmarks}
Various prompts were updated and standardized compared to evaluations in \citet{sellergren2025medgemma}. In particular, for similar tasks, like MCQ, we now now utilize the same initial instruction prompt to be more consistent. Updated prompts are shown in Appendix~\ref{sec:medgemma1.5_prompts}. 
Similar to the previous MedGemma models, we manually optimized prompts for MedGemma 1.5 on the training and validation splits to find prompts that worked well for evaluations.
We briefly summarize the previous evaluation datasets here. See~\citet{sellergren2025medgemma} for further details. 

For chest X-ray classification, performance was measured via F1-score across three datasets: MIMIC-CXR \cite{johnson2019mimic, johnson2019mimicdatabase, goldberger2000physiobank}, CheXpert~\cite{irvin2019chexpert}, and the ChestX-ray14 (CXR14)~\cite{wang2017chestx} dataset.
On MIMIC-CXR, we evaluated five common lung conditions (atelectasis, cardiomegaly, consolidation, edema, and pleural effusion). 
Two test sets were used for MIMIC-CXR: a Med-Gemini Test Set~\cite{yang2024advancing}, using radiologist-adjudicated labels instead of the original ones, and treating only explicitly 0-labeled conditions as negatives, i.e. leaving out non-mentioned or uncertain conditions, and a MAIRA test set using case selection from \citet{hyland2023maira} and original labels from \citet{MIMIC-CXR-JPG2019} (treating uncertainty as negative). 
On ChestX-ray14, we evaluated three conditions--lung opacity, pneumothorax, and fracture--on radiologist-adjudicated labels from \citet{majkowska2020chest}. For CheXpert, we evaluated on all 14 labeled observations \cite{irvin2019chexpert} using the original labels.

For image-based MCQ, we evaluated accuracy on DermMCQA~\cite{liu2020deep}, PathMCQA \cite{sellergren2025medgemma, jaroensri2022deep,nagpal2019development,nagpal2020development, sadhwani2021comparative}, and EyePACS \cite{cuadros2009eyepacs}.
DermMCQA~\cite{liu2020deep} is an internal dermatology dataset consisting of one image per patient from 1996 patients. There are 136 different skin conditions in total, with ground truth diagnoses provided by dermatologists based on the images and metadata. We generated 4-option multiple choice questions (MCQs) using random distractors for each image.
PathMCQA is an internal histopathology dataset, with a test split comprising of 450 patches from 354 whole slide images across breast, lung, prostate, lymph, and cervical specimens. Identification, grading, and sub-typing tasks were formulated as 4--9 option MCQs with ground truth from board-certified pathologists~\cite{sellergren2025medgemma}. On the EyePACS fundus image dataset~\cite{cuadros2009eyepacs}, we evaluated against clinically determined, 5-class diabetic retinopathy severity grades, formatted as 5-option MCQs, for 3614 de-identified images, each originating from a different patient.

For general visual question answering of medical images, we evaluated average tokenized F1 on SLAKE \cite{liu2021slake} and VQA-RAD \cite{lau2018dataset}. SLAKE is a dataset of aggregated CT, MRI, and X-Ray images and questions of body parts (neck, pelvis, abdomen, head and chest). We used the default splits for SLAKE. VQA-RAD is a dataset of radiology images and questions of head, chest, and abdomen. For VQA-RAD, we used splits from \citet{yang2024advancing} instead of the original splits in order to prevent contamination of the test split.

We also reported a radiology report generation metric (RadGraph F1~\cite{jain2021radgraph}) on the 912-image MIMIC-CXR test set used in~\citet{tanno2024consensus} and \citet{yang2024advancing}.

For general medical text MCQ, we also evaluated accuracy on the publicly available test splits for MedQA, MedMCQA, PubMedQA, MMLU medical subcategories, AfriMed-QA, and MedXpertQA \cite{zuo2025medxpertqa}. Note that MedXpertQA is out of distribution.

\subsection{New Multi-Modal Evaluations}

\subsubsection{Condition Classification from 3D CT and MR Images} 

\begin{figure}[htp] \centering \includegraphics[width=1\linewidth]{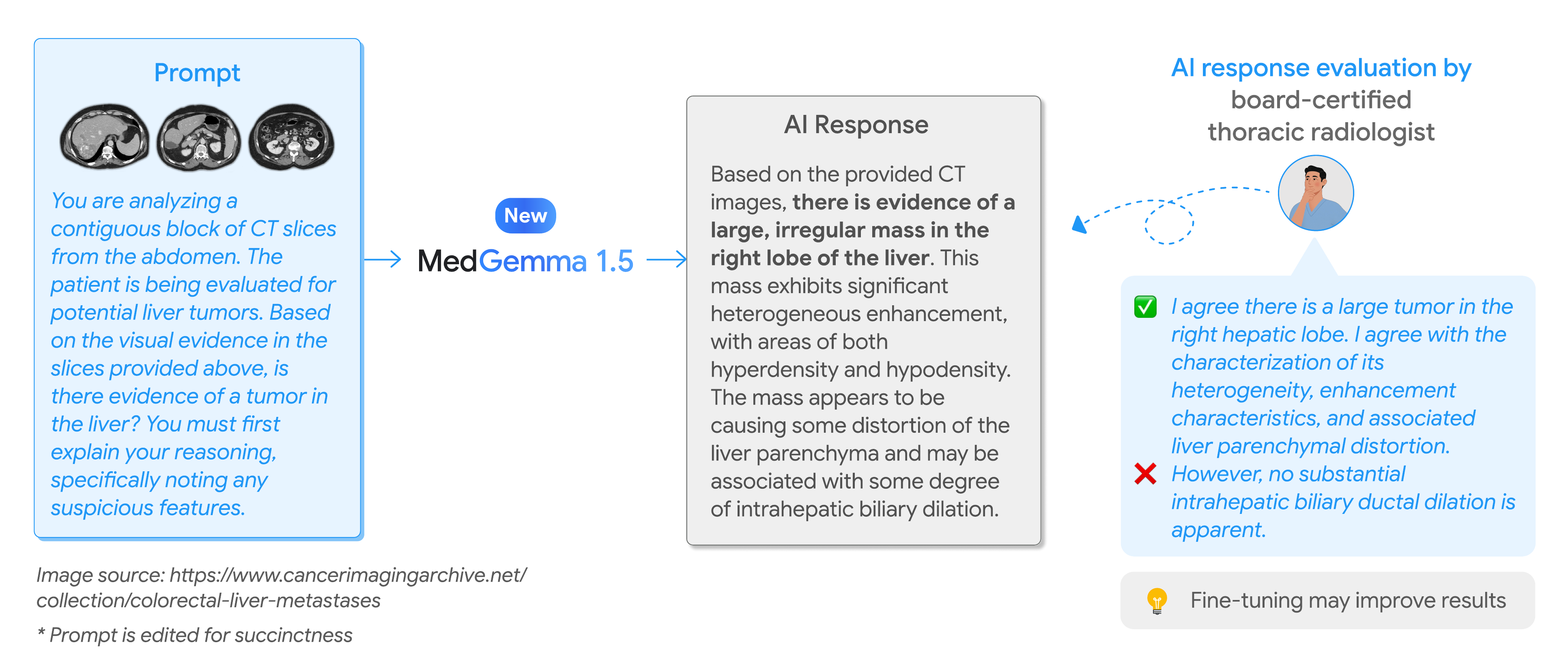} \caption{\small{\textbf{Overview of MedGemma 1.5 capabilities.} The updated 4B architecture now supports 3D radiology (CT/MRI volumes).}} \label{fig:3d_CT} \end{figure}

On CT and MR images we evaluated MedGemma and comparable models on the classification of common conditions.

The test split of the internal CT Dataset 1 consisted of head, chest, and abdominal/pelvis acquisitions, Models were evaluated on their ability to detect the following conditions: cardiac calcification, suspicious lung nodules (chest), aortic aneurysm, renal calculus, tumors, appendicitis (abdomen/pelvis), and hemorrhage (head). To extract binary ground truth labels (indicating presence or absence within the volume), we employed a multi-stage pipeline: (1) RegEx-based screening for positive mentions, (2) Gemini-based extraction, and (3) final manual review of a random subset by a US board-certified cardiothoracic radiologist. A condition was labeled 'absent' if the report explicitly ruled it out or omitted any mention of it.
The MRI Dataset 1 test split comprised brain, knee, and abdomen acquisitions. Labels were extracted using the same pipeline for the following conditions: acute infarct, hemorrhage, and multiple sclerosis (brain); meniscal tears and fractures (knee); and liver disease and pancreatic lesions (abdomen).
Both internal datasets were sub-sampled without replacement to ensure a balanced distribution of positive and negative cases per condition. Image volumes were preprocessed as described in Section~\ref{sec:3d_ct}. 

We also evaluated CT performance on the (internal) validation split of the public CT-RATE dataset~\cite{hamamci2024developing}, comprising of 1,564 non-contrast chest CT acquisitions from 1304 patients. As with the internal dataset, we measured macro accuracy across the binary prediction of 18 different conditions and abnormalities, respectively, while using the original labels provided with the dataset, which mostly cover common lung and cardiology conditions as described in~\cite{hamamci2024developing}. We preprocessed CT volumes in the same manner as described above, although not starting with raw volumes, but ones resampled to $480 \times 480 \times 240$, as provided by~\cite{hamamci2024developing}.

During evaluation, models were queried for each condition separately. Prompts consisted of the sequence of selected image slices interleaved with slice indices (e.g., \texttt{SLICE \{index\}}), followed by a binary question regarding the condition's presence. For the internal datasets, the prompt also included the patient history from the original report (see Table~\ref{tab:3d_ct_prompt}).
Because the models produced generative text rather than class probabilities, evaluation was based on binary presence/absence answers. We first computed performance metrics for each condition independently—using Accuracy for the balanced internal sets and F1 score for the imbalanced CT-RATE dataset—before calculating the Macro-Accuracy and Macro-F1 scores, respectively, to provide an aggregate measure of model performance. Note that CT-RATE results are included in Appendix~\ref{app:CT-RATE}.


\subsubsection{Pathology Report Generation from WSI} 

We evaluated using the ROUGE metric against the final diagnosis sections of the original pathology reports, having previously showed that this metric provides good correlation with pathologist scoring of image-text pairs for the same task \cite{ahmed2025polypath}. Each WSI was processed and preprocessed (with the patch extraction methods specified in Section~\ref{sec:wsi}) and provided as input along with the text specimen label (e.g. ``colon biopsy''). For this analysis, only single WSI-text pairs are used. 
\subsubsection{Longitudinal Chest X-Ray}
To assesses the model's capacity for temporal reasoning within longitudinal medical imaging, we evaluated on MS-CXR-T \cite{PhysioNet-ms-cxr-t-1.0.0, bannur2023learning, physionet}. This task required the model to analyze pairs of chest X-rays—specifically a prior and a current study—to determine the trajectory of five specific cardiopulmonary pathologies: consolidation, edema, pleural effusion, pneumonia, and pneumothorax.
The evaluation pipeline combined two radiographs with a structured text prompt, where for every image pair, the model is prompted to return one of three potential classes: (A) Improved, (B) Stable, or (C) Worsened. 
Performance is quantified using the macro accuracy metric to account for class imbalance \cite{bannur2023learning}.

\subsubsection{Localization of Anatomical Regions}
To assess the model's ability to localize anatomical regions of interest, we evaluated on Chest ImaGenome \cite{wu2021chest}. We utilized a bounding box evaluation protocol. The model was prompted to generate coordinates of 2D bounding boxes for specific anatomical structures identified in the query. 
The primary metric for evaluating localization performance was the Intersection over Union (IoU). For a predicted bounding box $B_{pred}$ and a ground truth bounding box $B_{gt}$, the IoU is defined as in \citet{wu2021chest}.





The model was provided with an input frontal chest X-ray image and a text prompt. The prompt contained specific instructions to output a JSON list of objects, where each object is defined by a label and a 2D bounding box coordinate set `[y0, x0, y1, x1]`. The coordinates were requested to be normalized to the range $[0, 1]$, with $(y0, x0)$ representing the top-left corner and $(y1, x1)$ the bottom-right corner.


\subsubsection{Document Understanding: Structured Data Extraction from Lab Reports}
MedGemma 1.5 was tuned and evaluated on structured data extraction from multimodal laboratory reports, converting key attributes from document images and PDFs (rendered into an image via an open-source library\footnote{\url{https://github.com/pypdfium2-team/pypdfium2}}) into a structured JSON format. This task may be considered a prerequisite for downstream standardization and interoperability tasks such as LOINC (Logical Observation Identifiers Names and Codes) mapping or FHIR (Fast Healthcare Interoperability Resources) resource generation.

To ensure robust generalization, we curated a composite dataset designed to reflect the complexities of real-world clinical data. The corpus inputs consist of medical laboratory reports provided as document images (e.g., PNG, JPEG) and Portable Document Formats (PDFs). The scope is specifically limited to pathology lab tests.
The datasets encompass both digital-native reports and scanned documents, the latter introducing challenges such as noise, variable illumination, and rotational artifacts inherent to manual digitization. 

The target JSON objects are structured to reflect the hierarchical and relational data of the source documents, focusing on the following data: name, result, unit, specimen, method, and sample collection time. F1 performance is calculated using a multi-phase label matcher algorithm to pair predicted labels with ground truth counterparts, followed by a metric calculation component that computes overall and granular (per-parameter) metrics: precision, recall, and F1 scores. 

The required output for this task is a structured JSON object that accurately reflects the hierarchical and relational data of the source document, maintaining accurate key-value pairings for all extracted entities.
This processing was applied to all EHR dataset 2, EHR dataset 3, EHR dataset 4, and Mendeley Clinical Laboratory Test Reports.

\subsection{New Text-based Evaluations}


\paragraph{EHRNoteQA:} 
To assess the model's capability in clinical reasoning over real-world electronic health records, we utilized the EHRNoteQA benchmark~\cite{kweon2024ehrnoteqa}. 
This dataset comprises 962 question-answer pairs derived from discharge summaries in the MIMIC-IV database, covering diverse clinical topics such as treatment plans, diagnostics, and patient history. 
Each instance involves analyzing accumulated discharge summaries for a specific patient to answer clinically relevant questions.

While the benchmark supports both open-ended and multiple-choice formats, our evaluation focused exclusively on the multiple-choice question (MCQ) setting. 
For each query, the model was presented with the patient's discharge notes, a question, and five answer choices (A through E) and overall accuracy was assessed.

%% file: sections/discussion.tex
The release of MedGemma 1.5 marks an important step for open-source medical artificial intelligence. Its performance across diverse benchmarks demonstrates that a single, efficient 4B parameter model can not only retain competency across established text-based benchmarks but also generalize to complex tasks in high dimensional and longitudinal imaging. 
Notably, due to the improved distillation and RL, we see that the model is able to actually improve performance on multiple text-based benchmarks compared to 1.0, including 5\% accuracy on MedQA and 8\% on MedXperQA (multimodal).
The model's ability to process native modalities with spatial and temporal depth—synthesizing evidence across non-contiguous volumetric scans, whole-slide digital pathology, and longitudinal imaging—demonstrates that smaller LLMs can make progress towards understanding of 3D anatomy and disease progression.

Beyond strong benchmark performance, these architectural improvements establish MedGemma 1.5 as a highly practical foundation for developers. Innovations like the enhanced anatomical localization and multi-timepoint analysis provide out-of-the-box spatiotemporal awareness, while its robust EHR and Lab Report parsing capabilities enable medical-specific OCR use-cases. Importantly, these out-of-the-box functionalities are designed as foundational data processing tools, distinct from automated clinical decision-making or the practice of medicine. Rather than engineering complex multimodal pipelines from scratch, developers and researchers can leverage this efficient baseline as a versatile starting point to fine-tune bespoke clinical applications, accelerating the deployment of accessible, next-generation healthcare tools.

\paragraph{Comparisons}
To contextualize MedGemma 1.5 within the broader ecosystem of open-weights models, we compared its performance against Qwen3 VL 4B, a similarly sized state-of-the-art multimodal model\footnote{Released about half a year after Gemma 3}. The comparison reveals distinct design philosophies. Qwen3 VL 4B demonstrates superior performance on general text-based biomedical knowledge tasks, such as MedQA. 
However, MedGemma 1.5 significantly outperforms Qwen3 VL 4B in specialized clinical vision tasks that require domain-specific visual capabilities. On all vision tasks, MedGemma 1.5 achieved higher performance. This divergence highlights the utility of MedGemma's specialized post-training and distillation pipeline; while generalist models excel at knowledge retrieval, the MedGemma lineage retains an edge in the interpretation of nuanced multimodal reasoning (e.g. via its superiority in multimodal MedXpertQA and all Medical image classification tasks). 

In our comparative analysis of novel evaluation tasks, we aimed to provide a broad perspective by including external baselines such as Qwen3 VL 4B. Some evaluations were not possible to perform with Qwen3 due to logistical constraints within our current internal evaluation framework. Future work may involve extending this framework to accommodate the distinct inference protocols of the Qwen architecture.

\paragraph{Limitations}
Compared to MedGemma 1, we observed trade-offs inherent to the expansion of model capabilities. As MedGemma 1.5 became more of a ``medical generalist'', we noted minor regressions in specific legacy benchmarks, such as SLAKE \cite{liu2021slake} and VQA-RAD \cite{lau2018dataset}. 
However, we note that these benchmarks have their own limitations in terms of quality as evaluations rely on token overlap, and the ground truth answers are not standardized. We believe this new model is more generally useful at medical imaging due to the improved performance on high-dimensional imaging and bounding box localization. Developers seeking maximum performance on narrower tasks can bridge these gaps through targeted fine-tuning.

%% file: sections/conclusion.tex
MedGemma 1.5 significantly expands the utility of the original model by advancing beyond standard 2D tasks to tackle high-dimensional, spatiotemporal modalities such as 3D radiology, pathology whole-slide imaging, and longitudinal imaging sequences. Despite these complex new capabilities and added document-understanding features, the model retains the computational and cost efficiency of a 4B parameter architecture. By releasing this highly capable, multimodally-aware foundation as an open resource, we aim to empower the developer community with an even more useful starting point to bridge the gap between academic benchmarks and impactful, real-world clinical applications.

%% file: sections/ack.tex
\subsubsection*{Contributions}

\begin{multicols*}{2}
\setlength{\columnsep}{5pt}
\small
\vspace{-1.0\baselineskip}
\begin{itemize}[leftmargin=1em,rightmargin=0em]
\setlength\itemsep{0pt}
    \item[] \textbf{Technical Leads} \\\vspace{-6pt}
    \item[] Andrew Sellergren\textsuperscript{*}
    \item[] Fereshteh Mahvar
\end{itemize}

\vspace{-1.0\baselineskip}
\begin{itemize}[leftmargin=1em,rightmargin=0em]
\setlength\itemsep{0pt}
    \item[] \textbf{Core contributors} \\\vspace{-6pt}
    \item[] Chufan Gao\textsuperscript{*} 
    \item[] Timo Kohlberger
    \item[] Fayaz Jamil
    \item[] Madeleine Traverse
    \item[] Alberto Tono
    \item[] Bashir Sadjad
    \item[] Lin Yang
    \item[] Charles Lau
    \item[] Liron Yatziv
    \item[] Tiffany Chen
    \item[] Bram Sterling
    \item[] Kenneth Philbrick
    \item[] Richa Tiwari
    \item[] Yun Liu
    \item[] Madhuram Jajoo
    \item[] Chandrashekar Sankarapu
    \item[] Swapnil Vispute
    \item[] Harshad Purandare
    \item[] Abhishek Bijay Mishra
\end{itemize}

\vspace{-1.0\baselineskip}
\begin{itemize}[leftmargin=1em,rightmargin=0em]
\setlength\itemsep{0pt}
    \item[] \textbf{Contributors} \\\vspace{-6pt}
    \item[] Sam Schmidgall
    \item[] Tao Tu
    \item[] Anil Palepu
    \item[] Chunjong Park
    \item[] Tim Strother
    \item[] Rahul Thapa
    \item[] Yong Cheng
    \item[] Preeti Singh
    \item[] Kat Black    
\end{itemize}


\vspace{-1.0\baselineskip}
\begin{itemize}[leftmargin=1em,rightmargin=0em]
\setlength\itemsep{0pt}
    \item[] \textbf{Sponsors} \\\vspace{-6pt}
    \item[] Yossi Matias
    \item[] Katherine Chou
    \item[] Avinatan Hassidim
    \item[] Kavi Goel
    \item[] Joelle Barral
    \item[] Tris Warkentin
\end{itemize}


\columnbreak

\vspace{-1.0\baselineskip}
\begin{itemize}[leftmargin=1em,rightmargin=0em]
\setlength\itemsep{0pt}
    \item[] \textbf{Leads} \\\vspace{-6pt}
    \item[] Shravya Shetty
    \item[] Dale Webster
    \item[] Sunny Virmani
    \item[] David F. Steiner
    \item[] Can Kirmizibayrak
    \item[] Daniel Golden\textsuperscript{$\dagger$}
    \item[] \footnotesize{$\dagger$ Last author}
    \item[] \footnotesize{$*$ Co-first author}
\end{itemize}

\end{multicols*}

\subsubsection*{Acknowledgements}
Many teams from both Google Research and Google DeepMind collaborated extensively on this project.
We thank Ellery Wulczyn and Greg Corrado for their feedback and insight, which significantly enhanced this report. 

\subsubsection*{Use of AI in Manuscript Preparation}
Sections of this manuscript was drafted using Gemini 2.5 Pro and Gemini 3 Pro and further refined via human editors over multiple rounds. Final manual checks were performed to ensure content accuracy. The authors take full responsibility for the content.

%% file: sections/appendix.tex
\section{CT-RATE Evaluation}
\label{app:CT-RATE}
We additionally evaluated a portion of our models on the CT-RATE dataset \cite{hamamci2024developing}, where we process accordingly (without resampling) per Section~\ref{sec:3d_ct} with results summarized in Table~\ref{tab:ctrate}. Unlike specialized, custom-built CT architectures that are optimized to yield multi-label predictions in a single forward pass, applying generalist vision-language models to this high-dimensional task required a more granular inference strategy. Specifically, our framework necessitated querying the model 18 times per condition to accurately parse the diagnostic signal. This iterative interrogation—while essential for leveraging the generative capabilities of the model—introduced a substantial computational bottleneck compared to the efficient, one-shot inference typical of dedicated CT classifiers. Consequently, due to the lengthy runtime associated with this eighteen-fold increase in query volume, we limited this evaluation to a representative subset of our models. We chose to evaluate on the most relevant models, as well as Gemini 3.0 Flash (which has shown to be competitive with Gemini 3.0 Pro in multimodal tasks).

The results presented in Table~\ref{tab:ctrate} highlight a significant divergence in performance between domain-specialized and generalist architectures on high-dimensional medical imaging tasks. 
Most notably, the MedGemma 4B models (versions 1 and 1.5) demonstrates superior zero-shot generalization capabilities compared to the general-purpose Gemini 3.0 Flash model. Despite the CT-RATE dataset being out-of-distribution (OOD) for the MedGemma training curriculum, these models achieved much higher Macro F1 scores. 
This disparity likely stems from the domain-specific pretraining of MedGemma, enabling a robust "medical prior," enabling more effective feature extraction from volumetric data even when the specific dataset distribution is novel. 

\begin{table}[thbp]
\centering
\caption{\small\textbf{CT Rate Results}}
\label{tab:ctrate}
\begin{tabular}{@{}l l c c c }
\toprule
Task & Metric & \makecell[c]{MedGemma\\1 4B} & \makecell[c]{MedGemma\\1.5 4B} & \makecell[c]{Gemini 3.0\\Flash} \\
\midrule

CT-RATE$^*$ (3D CT) & Macro F1 & 23.5 & 26.9 & 8.5 \\
\bottomrule
\end{tabular}
\end{table}


\begin{table}[ht]
\centering
\caption{\small{\textbf{Accuracy results on general, non-medical benchmarks.}}}
\label{tab:model_comparison_mmlu}
\resizebox{\linewidth}{!}{
\begin{tabular}{llccccc}
\toprule
Type & Benchmark  & \makecell{MedGemma 1 4B}&\makecell{MedGemma 1.5 4B}& \makecell{Gemma 3 4B\textsuperscript{\S}}  & \makecell{MedGemma 1 27B}& \makecell{Gemma 3 27B\textsuperscript{\S}}\\
\midrule
\multirow{1}{*}{Text-only} & MMLU Pro  &  39.1 &  33.8 & 43.6 & 60.2 & \textbf{67.5} \\
\bottomrule
\end{tabular}
}
\begin{tablenotes}
  \scriptsize
  \item $\S$ Prior reported results
\end{tablenotes}\end{table}

Table~\ref{tab:model_comparison_mmlu} shows a degradation in general knowledge reasoning compared to both its predecessor, MedGemma 1 4B and the foundational Gemma 3 4B. This decline suggests a clear tradeoff in the 4B parameter class, indicating that the intensive fine-tuning required to specialize the model for imaging may have resulted a diminished capacity for out-of-domain, general-purpose tasks. Still, we believe that this trade off is worth it for the large performance gains in the multimodal medical domain.

\section{Prompts}
\label{sec:medgemma1.5_prompts}
This section contains a complete list of all prompts for MedGemma 1.5. Prompts not listed here, such as prompts for running EHRQA, are identical to those used in \citet{sellergren2025medgemma}.

Please note that MedGemma models do not have a system instruction (system prompt). For general models, we apply the following system instruction to 
all evaluations that require radiology or chest X-ray interpretation: MS CXRT, SlakeVQA, VQA-Rad, Chest ImaGenome (Localization) \texttt{"You are a helpful radiology assistant."}. For all other benchmarks, we apply the following \texttt{"You are a helpful medical assistant."}.

For MedGemma 1.5, we also turn thinking on for certain benchmarks to encourage reasoning. Specificaly, we turn thinking on by appending \texttt{"SYSTEM INSTRUCTION: think silently if needed."} to the system prompt for the following benchmarks: MedQA, MedMCQA, EHRNoteQA, PubMedQA, MMLU Med, MedXpertQA (Text Only), and AfriMed-QA.

\begin{table}[ht]
\centering
\caption{Updated Prompts (Version 1.5) for General Medical Text Evaluation Tasks}
\label{tab:text_prompts_v1.5}
\begin{tabular}{|p{0.25\linewidth}|p{0.75\linewidth}|}
\hline
\textbf{Task} & \textbf{Prompt Template / Suffix} \\ \hline
MedQA & \multirow{6}{=}{\texttt{\{\{ question \}\}\textbackslash nYou may write out your argument before stating your final, very short, definitive, and concise answer (no more than a few words or the letter corresponding to your answer choice if the question is multiple choice) X in the format "Final Answer: X":}} \\ 
PubMedQA & \\
MedMCQA & \\
MMLU & \\
MedXpertQA (Text) & \\
AfriMed & \\ \hline
\end{tabular}
\end{table}

\begin{table}[ht]
\centering
\caption{Updated Prompts (Version 1.5) for Multimodal Binarized MCQ Evaluation Tasks. These tasks involve binary classification (Yes/No) and use a strict format constraint in Version 1.5.
}
\label{tab:binarized_prompts_v1.5}
\begin{tabular}{|p{0.2\linewidth}|p{0.75\linewidth}|}
\hline
\textbf{Task} & \textbf{Prompt Template} \\ \hline
MIMIC-CXR & \multirow{3}{=}{\texttt{\{\{ image \}\} + \{\{ question \}\} You MUST end your responce with either "Final Answer: yes" or "Final Answer: no}} \\ 
CheXpert &  \\ 
CXR14 &  \\ \hline
\end{tabular}
\end{table}

\begin{table}[ht]
\centering
\caption{Updated Prompts (Version 1.5) for Visual Evaluation Tasks. These tasks cover general visual question answering and specific diagnostic classification.}
\label{tab:visual_prompts_v1.5}
\begin{tabular}{|p{0.2\textwidth}|p{0.75\textwidth}|}
\hline
\textbf{Task} & \textbf{Prompt Template} \\ \hline
SlakeVQA & \texttt{\{\{ image \}\} + \{\{ question \}\} You may write out your argument before stating your final, very short, definitive, and concise answer X (no more than a few words or the letter corresponding to your answer choice if the question is multiple choice) in the format "Final Answer: X":} \\ \hline
VQA-Rad & \texttt{\{\{ image \}\} + \{\{ question \}\} You may write out your argument before stating your final, very short, definitive, and concise answer X (no more than a few words or the letter corresponding to your answer choice if the question is multiple choice) in the format "Final Answer: X":} \\ \hline
Pathology WSI & \texttt{\{\{ images \}\} + Provide a brief diagnostic text for the set of pathology patches extracted from a pathology slide. Consider the tissue type and procedure (below) when deciding what to include in the diagnostic text. \{\{ type\_procedure \}\} \{\{ question \}\}} \\ \hline
DermMCQA & \texttt{\{\{ image \}\} + \{\{ question \}\} You must choose the most likely diagnosis and respond with "The most likely diagnosis is:" followed by your choice letter.} \\ \hline
EyePACS & \texttt{\{\{ image \}\} + Given this fundus image, determine the most likely diabetic retinopathy (DR) stage present, even if you are unsure:\newline
      A: No DR\newline
      B: mild DR\newline
      C: moderate DR\newline
      D: severe DR\newline
      E: proliferative DR\newline
  You must choose the most likely diagnosis and respond with "The most likely diagnosis is:" followed by your choice letter. } \\ \hline
\end{tabular}
\end{table}

\begin{table}[ht]
\centering
\caption{Prompt (Version 1.5) for EHRNoteQA}
\label{tab:ehrnoteqa_prompt}
\begin{tabular}{|p{0.2\textwidth}|p{0.75\textwidth}|}
\hline
\textbf{Task} & \textbf{Prompt Template} \\ \hline
EHRNoteQA & \texttt{BEGIN\_INSTRUCTIONS\newline
Given the following discharge note for a patient, answer the question by only picking one of the A, B, C, D, E options. Each discharge note starts with "DISCHARGE ?:", the question starts with "QUESTION:" and the choices with "CHOICE\_?:" where ? is a single character. To answer, describe your thought process for each choice; finish your answer with "Final Answer: (?)" where ? is a single character indicating the correct choice.\newline
END\_INSTRUCTIONS\newline
\newline
\{discharge\_note\}\newline
\newline
QUESTION:\newline
\{orig\_question\}\newline
the choices are:\newline
CHOICE\_A: \{choice\_A\}\newline
CHOICE\_B: \{choice\_B\}\newline
CHOICE\_C: \{choice\_C\}\newline
CHOICE\_D: \{choice\_D\}\newline
CHOICE\_E: \{choice\_E\}\newline
Describe your thought process for each choice and end your answer with "Final Answer: (?)" where ? is a single character indicating the correct answer. Use this exact format at the end "Final Answer: (?)".} \\ \hline
\end{tabular}
\end{table}

\begin{table}[ht]
\centering
\caption{Prompt (Version 1.5) for Document Understanding PNG/PDF images to JSON}
\label{tab:docunderstanding}
\begin{tabular}{|p{0.25\textwidth}|p{0.7\textwidth}|}
\hline
\textbf{Task} & \textbf{Prompt Template} \\ \hline
EHR Dataset 2 & \multirow{8}{=}{\texttt{You are a Clinical Data Extraction Specialist. Your job is to parse lab reports with high precision.\newline
From the given lab report, extract all lab tests into a JSON list.\newline
Each test object in the list must include: name, result, unit, range, panel, method, specimen, sample\_collection\_time (formatted as DD-MM-YYYY HH:MM:SS)}} \\ 
EHR Dataset 3 &  \\ 
EHR Dataset 4 &  \\ 
Mendeley Clinical Laboratory Test Reports & \\ 
 & \\ 
 & \\ 
 & \\ 
\hline
\end{tabular}
\end{table}

\begin{table}[ht]
\centering
\caption{Prompt (Version 1.5) for Anatomy Localization Tasks}
\label{tab:bbox_prompt}
\begin{tabular}{|p{0.3\textwidth}|p{0.65\textwidth}|}
\hline
\textbf{Task} & \textbf{Prompt Template} \\ \hline
Chest ImaGenome (Localization) & \texttt{\{\{ image \}\} + Where is the \{object\}?} \\ \hline
\end{tabular}
\end{table}

\begin{table}[ht]
\centering
\caption{Prompt (Version 1.5) for 3D CT Volumetric Analysis}
\label{tab:3d_ct_prompt}
\begin{tabular}{|p{0.3\textwidth}|p{0.65\textwidth}|}
\hline
\textbf{Task} & \textbf{Prompt Template} \\ \hline
CT-US1 & \texttt{\{\{ images \}\} + After looking at the indication and patient history "\{HISTORY\}"... Is there "\{Label\}" in the CT volume? You may write out your argument before stating your final answer "Final Answer: yes" or "Final Answer: no".} \\ \hline
MRI-US1 & \texttt{\{\{ images \}\} + 'After looking at the patient history "{history\_text}", Is there \{\{label\}\} in the MRI volume? You may write out your argument before stating your final answer ""Final Answer: yes"" or ""Final Answer: no""} \\ \hline
CT-RATE & \texttt{\{\{ images \}\} + You are an expert radiologist for chest CT. Looking at these CT slices, is there Emphysema? Answer with 'Final Answer: yes' or 'Final Answer: no'} \\ \hline
\end{tabular}
\end{table}